%% file: main.tex
\documentclass[lettersize,journal]{IEEEtran}
\usepackage{amsmath,amsfonts}
\usepackage{array}
\usepackage[caption=false,font=normalsize,labelfont=sf,textfont=sf]{subfig}
\usepackage{textcomp}
\usepackage{stfloats}
\usepackage{url}
\usepackage{verbatim}
\usepackage{graphicx}
\usepackage{cite}
\usepackage{multirow}
\usepackage{booktabs}
\usepackage{amsfonts}
\usepackage{amsmath}
\usepackage{color,xcolor}
\usepackage{diagbox}
\usepackage{bbm}
\usepackage[backref]{hyperref} 
\usepackage[utf8]{inputenc}
\usepackage[square,sort,comma,numbers]{natbib}
\usepackage{pifont}
\usepackage{picins}
\usepackage{multirow}
\usepackage{array}
\usepackage{algorithm}
\usepackage{algpseudocode}
\usepackage{listings}
\usepackage{subcaption}
\usepackage{setspace} 
\usepackage{graphicx}
\usepackage{subcaption}
\usepackage{caption}

\newcolumntype{L}[1]{>{\raggedright\let\newline\\\arraybackslash\hspace{0pt}}m{#1}}
\newcolumntype{C}[1]{>{\centering\let\newline\\\arraybackslash\hspace{0pt}}m{#1}}
\newcolumntype{R}[1]{>{\raggedleft\let\newline\\\arraybackslash\hspace{0pt}}m{#1}}

\begin{document}

\title{\raisebox{-0.15cm}{\includegraphics[height=0.7cm]{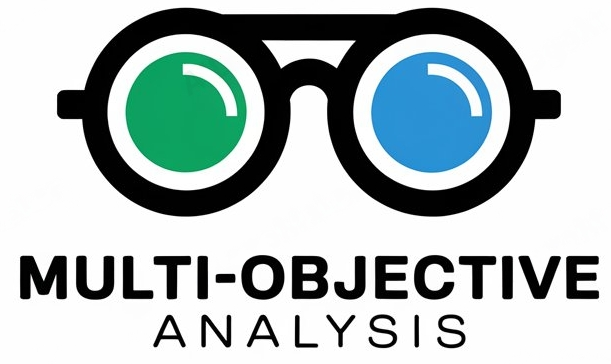}}  MOSABench: Multi-Object Sentiment Analysis Benchmark for Evaluating MLLMs}

\author{
    \IEEEauthorblockN{Shezheng Song, Chengxiang He, Shan Zhao, Chengyu Wang} \\
    \IEEEauthorblockN{Qian Wan, Tianwei Yan, Meng Wang, \textit{IEEE Fellow}} \\
    \thanks{Shezheng Song is with the College of Computer, National University of Defense Technology, Changsha 410073, China and also with the School of Computer and Information Engineering, Hefei University of Technology, Hefei 230009, China (e-mail: \{ssz614@nudt.edu.cn).}
    \thanks{Chengxiang He, Shan Zhao, Meng Wang is with the School of Computer and Information Engineering, Hefei University of Technology, Hefei 230009, China (e-mail: zhaoshan@hfut.edu.cn, eric.mengwang@gmail.com).}
    \thanks{Qian Wan is currently an Assistant Researcher with the CCNU, Wuhan, China.}
    
}

\markboth{IEEE TRANSACTIONS ON Multimedia, VOL. XX, NO. XX, DECEMBER 2023}%
{Shell \MakeLowercase{\textit{et al.}}: A Sample Article Using IEEEtran.cls for IEEE Journals}

\IEEEpubid{0000--0000/00\$00.00~\copyright~2021 IEEE}


\maketitle

\begin{abstract}
    Multimodal large language models (MLLMs) have shown remarkable progress in high-level semantic tasks such as visual question answering, image captioning, and emotion recognition. However, despite advancements, there remains a lack of standardized benchmarks for evaluating MLLMs performance in multi-object sentiment analysis, a key task in semantic understanding. To address this gap, we introduce MOSABench, a novel evaluation dataset designed specifically for multi-object sentiment analysis. MOSABench includes approximately 1,000 images with multiple objects, requiring MLLMs to independently assess the sentiment of each object, thereby reflecting real-world complexities. Key innovations in MOSABench include distance-based target annotation, post-processing for evaluation to standardize outputs, and an improved scoring mechanism. Our experiments reveal notable limitations in current MLLMs: while some models, like mPLUG-owl and Qwen-VL2, demonstrate effective attention to sentiment-relevant features, others exhibit scattered focus and performance declines, especially as the spatial distance between objects increases. This research underscores the need for MLLMs to enhance accuracy in complex, multi-object sentiment analysis tasks and establishes MOSABench as a foundational tool for advancing sentiment analysis capabilities in MLLMs.
\end{abstract}


\section{Introduction}
\label{sec:intro}

\begin{figure}[htbp]
    \centering
    \includegraphics[width=\linewidth]{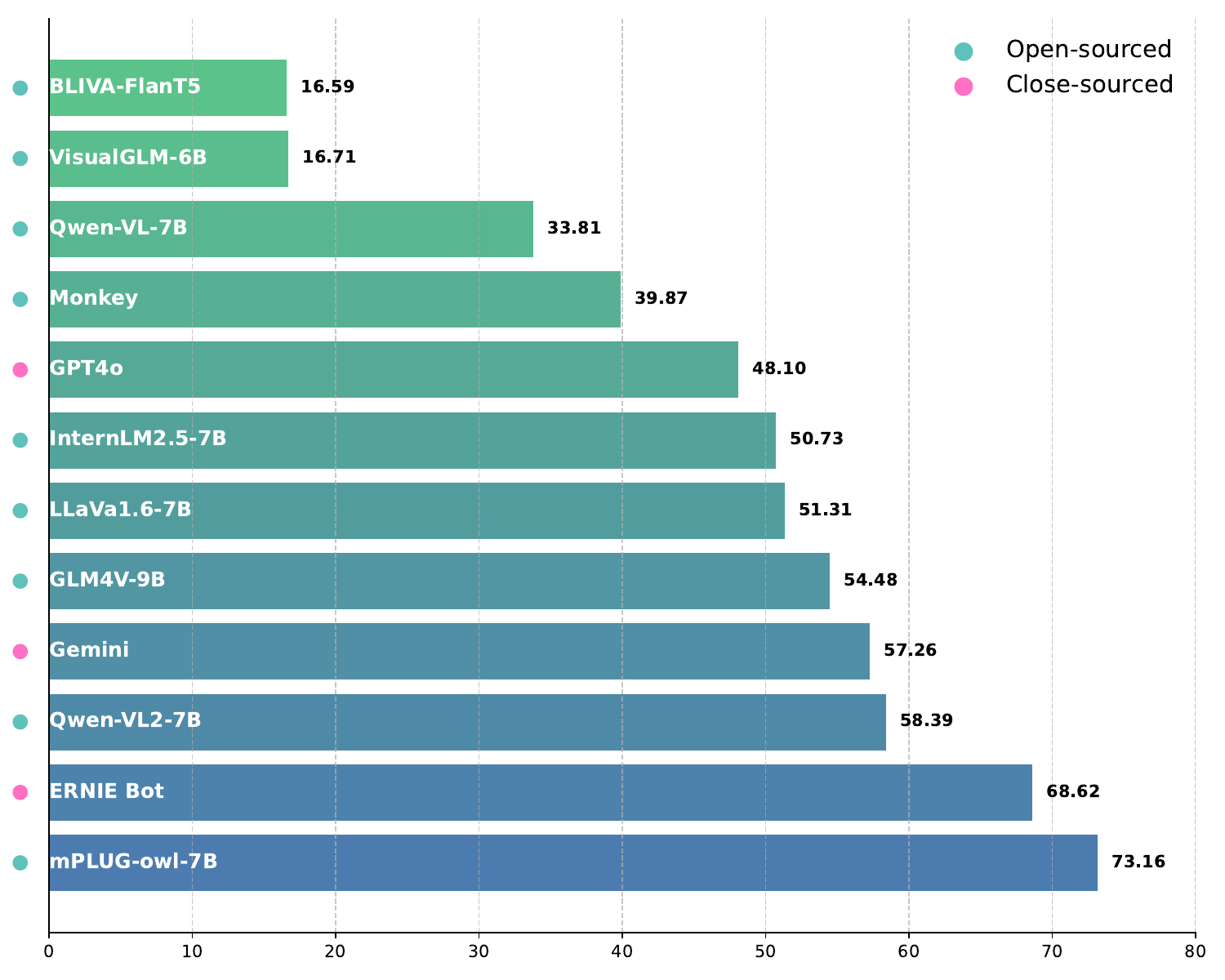}
    \caption{F1 comparison on MOSABench across multimodal large language models}
    \label{fig:F1}
\end{figure}

In recent years, multimodal large language models~\cite{yin2024survey} (MLLMs) have made significant progress in image understanding tasks, demonstrating immense potential, particularly in high-level semantic tasks such as visual question answering~\cite{antol2015vqa, wu2017visual}, image captioning~\cite{hossain2019comprehensive}, and emotion recognition~\cite{dzedzickis2020human, TMMmethod1, TMMmethod2, TMMmethod3}. Commercial platforms like GPT4o~\cite{GPT4} and Gemini~\cite{geminiteam2024geminifamilyhighlycapable}, as well as open-source models such as Flamingo~\cite{Flamingo} and LLaVA~\cite{llava}, have achieved outstanding performance on a wide range of traditional tasks, even surpassing human-level performance in tasks like multimodal named entity recognition~\cite{jia2023mner, ji2024cmner, TCSVT_ERC}. This impressive potential has driven the development of multimodal evaluation tasks. However, sentiment analysis~\cite{das2023multimodal, kaur2022multimodal}, as one of the key tasks in semantic understanding, still lacks a standardized benchmark specifically designed for evaluating MLLMs. Figure \ref{fig:F1} shows that the performance of MLLMs on our MOSABench is unsatisfactory.

\IEEEpubidadjcol

\begin{figure}[ht]
    \centering
    \includegraphics[width=\linewidth]{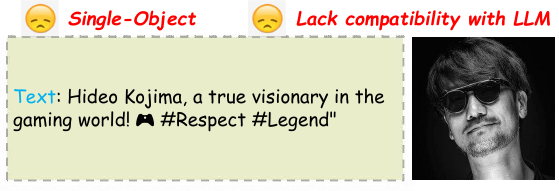}
    \caption{Example from a previous dataset, illustrating single-object data with a lack of instruction adaptation for MLLM.}
    \label{fig:pre-deficiency1}
\end{figure}
\begin{figure}[ht]
    \centering
    \includegraphics[width=\linewidth]{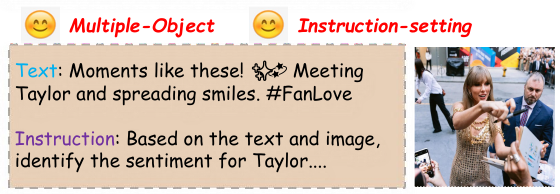}
    \caption{Example from our MOSABench.}
    \label{fig:pre-deficiency2}
\end{figure}

As shown in Figure \ref{fig:pre-deficiency1} and \ref{fig:pre-deficiency2}, current sentiment analysis datasets have limitations in accurately evaluating the capabilities of MLLMs.
\textbf{Lack of focus on multi-object understanding:}
In real-world social media, images typically contain multiple objects (such as people), each of which may convey different emotional information. 
Therefore, MLLMs are facing the challenge of multiple object sentiment analysis. However, most existing sentiment analysis benchmarks, such as Twitter15, Twitter17, and MSED~\cite{jia2022beyond}, are dominated by single-object samples, which leads to models focusing primarily on single-object sentiment classification ability~\cite{phan2019sentiment, TCSVT2023mcl}. When single-object data dominate the dataset, the model's good performance likely reflects its ability to classify emotions for individual objects, rather than its capacity to analyze emotions across multiple objects in an image. As a result, evaluating models on such imbalanced datasets fails to comprehensively assess their ability to understand multiple-object emotions, which limits the potential of multimodal sentiment analysis models. 
\textbf{Lack of adaptability to MLLMs}: Moreover, existing datasets~\cite{twitterdataset, jia2022beyond} have substantial limitations in evaluating the capabilities of MLLMs. Most of these datasets are designed for smaller models, lack instruction, and require extensive adaptation, making it challenging to comprehensively assess the performance of MLLMs. Additionally, since MLLMs generate outputs in varying formats rather than strictly following predefined formats, accurately evaluating their performance is more difficult.
Therefore, developing a standardized, multidimensional benchmark specifically designed for evaluating multi-object sentiment analysis tasks has become an urgent challenge.

To bridge this research gap, we introduce a new evaluation benchmark, MOSABench, comprising a dataset for testing and a scoring system adapted for MLLMs, offering a standardized and thorough assessment in multi-object sentiment analysis tasks. MOSABench contains approximately 1,000 text-image pairs, each with multiple objects, requiring the MLLM to independently assess the sentiment for each object to handle complex multi-object scenarios. Besides, MOSABench presents a standardized approach for assessing large model outputs in multi-object sentiment analysis. By integrating an enhanced scoring mechanism with F1 scores, we provide a comprehensive evaluation of model performance on multiple objects, advancing research and applications in this domain.
Specifically, MOSABench introduces three key innovations to advance the evaluation of multi-object sentiment analysis for MLLMs: distance-based object annotation, post-processing for evaluation, and improved scoring mechanism.
In \textbf{(1) distance-based target annotation}, we label the spatial distance between objects within images, revealing a significant relationship between object proximity and sentiment prediction accuracy—the further apart the objects are, the lower the accuracy tends to be. This provides a new perspective on MLLM limitations, highlighting that sentiment prediction accuracy can be influenced by the spatial arrangement of targets in images.
To address issues with inconsistent output formats in MLLM responses, we propose \textbf{(2) post-processing for evaluation}. This step is designed to standardize model outputs, reducing the impact of the format of MLLM responses on evaluation accuracy. By minimizing the influence of non-standard formats, our post-processing enables a more accurate assessment of MLLM capabilities in understanding image content.
Our \textbf{(3) improved scoring mechanism} introduces a unique multi-object evaluation approach. For each sample, separate sentiment judgments are required for two objects, assigning 3 points if both judgments are correct, 1 point if only one is correct, and 0 if both are incorrect. Together with traditional metrics like F1, Precision, and Recall, this scoring framework provides a comprehensive evaluation of model performance in handling multi-object sentiment tasks, allowing for a nuanced analysis of MLLM strengths and limitations.


We conduct a comprehensive evaluation of mainstream MLLMs on our MOSABench. Our analysis leads to the following key observations:
(1) \textbf{Challenges in MLLMs for multiple objects analysis}: Most current models perform suboptimally on the MOSABench multi-object sentiment analysis task, revealing significant challenges in handling multiple target emotions simultaneously. Although these models perform well on other tasks, such as VQA~\cite{antol2015vqa, goyal2017VQA} and MNER~\cite{jia2023mner, ji2024cmner}, they have limitations when it comes to multi-object sentiment analysis, which demands higher comprehension and reasoning abilities for multiple objects.
(2) \textbf{Impact of object distance on accuracy}: Our further analysis reveals a significant correlation between task accuracy and the spatial relationship between targets in the image. We categorize the distance between objects in the data into three types: Interlap, Close, and Far. The experimental results show that as the distance between targets increases, the performance of most MLLMs decreases. This finding provides a new perspective for improving model performance in multi-object sentiment analysis and emphasizes the importance of spatial relationships in multi-object contexts.
(3) \textbf{Performance variations across models}: We also observe significant performance differences across models. For example, the mPLUG-owl~\cite{mPLUG-Owl} and Qwen-VL2~\cite{Qwen2VL} models demonstrate relatively high stability and accuracy across various evaluation metrics. This may be due to their optimized architectures and larger parameter scales, which enable them to better adapt to complex sentiment analysis tasks. In contrast, models such as Qwen-VL and BLIVA-Flant5 show poorer performance, suggesting a lack of sufficient generalization capability in multi-object sentiment analysis, making it difficult for these models to accurately capture emotional features of multiple objects.

\section{Related Work}
\label{sec:related_work}



\subsection{Advances in MLLMs for Sentiment Analysis and the Demand for Scientific Benchmarking}
The recent advancements in large language models (LLMs)~\cite{Minigpt-4, shikra, wei2023lenna, IdealGPT} have significantly improved multimodal sentiment analysis, effectively handling complex sentiment tasks across various modalities~\cite{li2023multimodal, MultiModal-GPT}. Studies have demonstrated that LLMs achieve high accuracy on standard datasets, such as Twitter15 and Twitter17~\cite{twitterdataset}, which are widely used to assess sentiment analysis capabilities by integrating text and image data to analyze social media posts.
For example, the WisdoM\cite{wang2024wisdom} framework leverages large vision-language models (LVLMs) to enhance sentiment analysis by incorporating contextual world knowledge from images and text, thus improving LLMs interpretability and performance~\cite{krugmann2024sentimentai}. Similarly, the PSL~\cite{psl} framework is a pipeline-based approach for aspect-based sentiment analysis, using small language models (SLMs)~\cite{DTCA} for aspect extraction and MLLMs for sentiment analysis. This structured guidance enables MLLMs to focus on relevant image regions, effectively addressing the complexities of multimodal sentiment tasks~\cite{psl}. These frameworks highlight the progress LLMs have made in multimodal alignment and sentiment understanding~\cite{phan2019sentiment, krugmann2024sentimentai}.
However, current benchmarks, like Twitter15 and Twitter17~\cite{twitterdataset}, reveal limitations in assessing MLLMs true multimodal comprehension capabilities. Primarily, these datasets often lack image-text consistency, where the targets mentioned in the text may not appear in the associated image, hindering accurate evaluations of MLLMs capacity to integrate visual context~\cite{krugmann2024sentimentai}. Additionally, these benchmarks are not equipped with LLM-specific instructions~\cite{DreamLLM}, making it challenging to assess the impact of different prompting methods on sentiment prediction~\cite{psl}. Lastly, they lack multi-object sentiment assessment, which is crucial for evaluating the ability of MLLMs to independently analyze sentiments towards multiple entities within a single post~\cite{psl, wang2024wisdom}. These gaps underscore the need for a scientifically designed benchmark that captures multimodal nuances, includes structured prompts tailored for LLM tasks, and offers multi-object sentiment assessment to comprehensively evaluate LLM performance in real-world multimodal sentiment applications.




\subsection{Limitations of Current Benchmarks in Addressing Sentiment Analysis Needs}
In the evaluation of LLMs, tasks such as Named Entity Recognition (MNER)~\cite{ji2024cmner, jia2023mner} have developed dedicated benchmarks to better measure model performance. However, sentiment analysis lacks a specialized benchmark tailored specifically for LLMs. While some comprehensive benchmarks, such as MM-SOC~\cite{jin2024mmsoc} and MM-BigBench~\cite{yang2023mmbench}, include multimodal tasks related to sentiment analysis, they still exhibit notable limitations in design and evaluation~\cite{guo2023evaluating, wang2023evaluation}, particularly in areas of image-text consistency, instruction design, and multi-object sentiment judgment~\cite{wang2024wisdom, psl}.
For instance, MM-SOC primarily targets multimodal tasks within social media environments, covering sentiment analysis, hate speech detection~\cite{pi2023detgpt}, and more, using the Memotion dataset~\cite{ramamoorthy2022memotion} for joint image-text emotion detection. However, MM-SOC does not specifically address image-text consistency, which limits the evaluation of whether all entities mentioned in the text are also represented in the image. Additionally, MM-SOC lacks LLM-specific instructions, restricting its utility in large-scale multimodal tasks that require instruction-following capabilities~\cite{psl}.
MM-BigBench, on the other hand, covers a broad range of multimodal comprehension tasks, including visual question answering and multimodal sentiment analysis (MSA), focusing on multimodal information fusion and deep understanding. However, MM-BigBench does not provide detailed evaluations for multi-object sentiment analysis and does not adequately emphasize image-text consistency, which is essential for accurately identifying the sentiments toward multiple entities mentioned in the text. Furthermore, this benchmark lacks instructions specifically designed for LLMs, making it difficult to assess how different prompt structures might affect performance.


In summary, while current benchmarks have made progress in multimodal sentiment analysis, there remains significant room for improvement in image-text consistency, multi-object sentiment assessment, and instruction design specifically for LLMs.

\section{MOSA Benchmark Construction}

\begin{figure}[htbp]
    \centering
    \includegraphics[width=\linewidth]{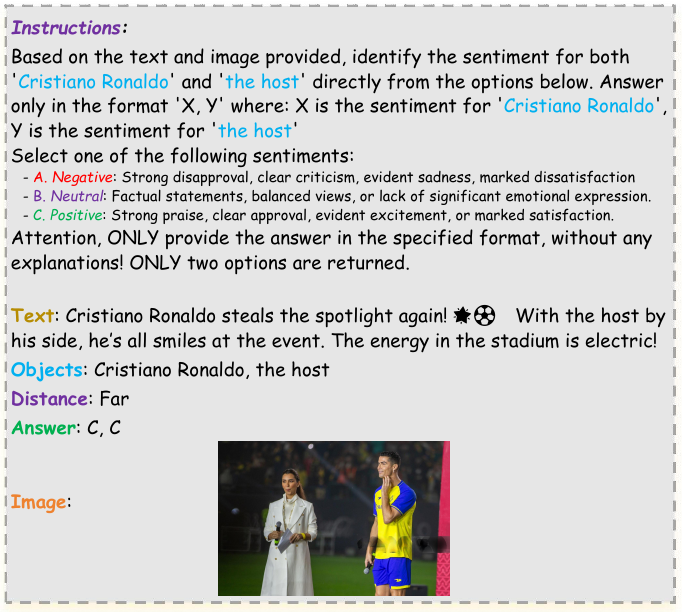}
    \caption{Data example of our MOSABench. ``Instruction'' specifies the task that the LLM needs to perform, while ``Answer'' represents the expected result of the task execution. ``Objects'' indicates the targets present in the image, requiring the LLM to complete the task in the ``Instruction'' by integrating the ``Text'' and ``Image''.
}
    \label{fig:data}
\end{figure}

\subsection{Dataset Construction}
We construct the MOSABench dataset to address limitations in current MLLMs for multi-object sentiment analysis tasks. First, samples are selected from Twitter15, Twitter17, and TwiBot-20 that meet the requirements for multi-object sentiment analysis. To ensure data accuracy and diversity, strict filtering criteria are applied: each text must contain multiple distinct targets, and each target must also appear in the corresponding image, enabling the model to capture emotional cues from both visual and textual information. Abstract terms such as “United Nations” are excluded to ensure that each target in the sample has a clear emotional expression, thereby enhancing the model in distinguishing sentiments across multiple targets.

During annotation, these samples are adapted from the original Multi-Aspect Based Sentiment Analysis (MABSA) format to a sentiment analysis format. Given a specified target, diverse question forms are designed, such as “Please confirm the sentiment of X” or “Please judge the status of X,” to simulate the model adaptability to different question types. This diverse design not only enhances the generalizability of the task but also prevents potential bias that may arise from a single question style, thereby enabling a more comprehensive evaluation of model capability.

To ensure objectivity and consistency in evaluation, the labeling system in our dataset is simplified by adopting a binary-choice structure in which “A” and “B” represent different sentiment categories, such as negative, neutral, and positive, using a fixed sentiment mapping. This structure reduces errors caused by inconsistent labels and streamlines model output processing, making sentiment analysis evaluation more straightforward. Additionally, the evaluation method is improved by implementing a novel scoring mechanism: for binary-choice questions, fully correct answers receive 3 points, partially correct answers receive 1 point, and fully incorrect answers receive no points. This scoring method enables a more comprehensive assessment of MLLM capabilities, enhancing the reliability and precision of sentiment analysis evaluation and providing a practical benchmark for multi-object sentiment analysis tasks.

\begin{figure*}[htbp]
    \centering
    \includegraphics[width=.8\linewidth]{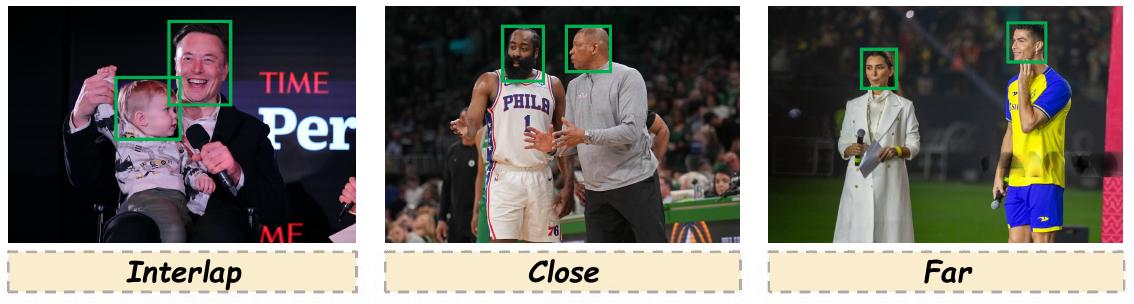}
    \caption{Distance label examples in MOSABench. ``Interlap'' indicates that the bounding boxes of the individuals overlap. ``Close'' denotes that the boxes do not overlap but the distance between them is less than $ L/k $, where $ L $ is the image length and $ k $ is a hyperparameter. ``Far'' signifies that the distance between the boxes exceeds $ L/k $.
}
    \label{fig:distance_example}
\end{figure*}

\input{table/3_alg}

\subsection{Distance Annotation}
As shown in Figure \ref{fig:distance_example}, we also annotate and analyze the spatial distances between targets within each image. This annotation strategy aims to verify the relationship between target distance and task accuracy, revealing the potential limitations of MLLMs in multi-object sentiment analysis. Experimental results show that target distance significantly impacts sentiment judgment accuracy; in particular, the accuracy decreases notably when targets are close to each other. This finding provides valuable insights for improving MLLM performance in complex scenarios.


As shown in this algorithm \ref{alg:distance_type}, we calculate the spatial relationship between two detected objects based on their bounding boxes. The bounding boxes, $B_1$ and $B_2$, are obtained using an object detection model, where only the two highest confidence detections of category "person" are selected as input. This ensures that the algorithm focuses on the spatial relation between two human figures within the image. The input image length $L$ and threshold parameter $k$ allow us to classify the relationship as \textit{Interlap}, \textit{Close}, or \textit{Far}. Specifically, the parameter $k$ serves as a tunable threshold to adjust the proximity level for determining \textit{Close} and \textit{Far} classifications.

\input{table/data_statistics}
\subsection{Datasets Statistics}
    We conduct a statistical analysis of our MOSABench dataset, with results presented in Table \ref{tab:dataset_statistics}. The dataset contains a total of 1,047 samples, categorized into three distance groups: \textit{Close}, \textit{Interlap}, and \textit{Far}, to capture varying spatial configurations. Among these samples, 57.11\% are classified as Close, 31.18\% as Interlap, and 11.71\% as Far, providing a balanced representation across different target proximities. This distribution facilitates a comprehensive evaluation of MLLM performance under diverse spatial contexts.
    
    The sentiment distribution in MOSABench reflects a scientifically consistent approach, aligning with the proportions observed in prior datasets. In MOSABench, Negative, Neutral, and Positive sentiments are represented by 15.44\%, 49.43\%, and 35.13\%, respectively. This design aligns closely with the sentiment distributions in the Twitter15 and Twitter17 datasets~\cite{twitterdataset}, which feature similar proportions: Twitter15 includes 12.06\% Negative, 59.29\% Neutral, and 28.65\% Positive samples, while Twitter17 comprises 12.19\% Negative, 45.68\% Neutral, and 42.13\% Positive samples. By mirroring these established distributions, MOSABench provides a scientifically robust benchmark for evaluating MLLMs sentiment analysis capabilities across multiple targets.

\subsection{MLLM Baselines}



Based on the MOSABench dataset we constructed, we conduct a comprehensive evaluation of existing MLLMs to assess their performance in multi-object sentiment analysis tasks. To this end, we propose two baseline methods and systematically test various mainstream model architectures, covering three distinct types.
In terms of model selection, we choose representative MLLMs: 
\begin{itemize}
    \item Open-sourced models: LLaVA1.6~\cite{llava}, mPLUG-Owl~\cite{mPLUG-Owl}, Qwen-VL~\cite{qwen_vl}, Qwen-VL2~\cite{Qwen2VL}, VisualGLM~\cite{du2022glm}, BLIVA-FlanT5~\cite{hu2023bliva}, Monkey~\cite{li2024monkey}, GLM4V~\cite{wang2023cogvlm}, InternLM2.5~\cite{InternLM2TechiReport}.
    \item Close-sourced models: ERNIE-Bot~\cite{sun2020ernie}, GPT-4o~\cite{GPT4} and Gemini~\cite{geminiteam2024geminifamilyhighlycapable}. 
\end{itemize}
This diverse selection of models ensures the broad applicability and representativeness of the evaluation results.


\subsection{Post-processing for LLM Scoring Assistance}

In our experiments, we observe that not all large language models generate outputs with consistent formatting. Some models show variability in response length or randomness due to pre-training biases, which makes it difficult to score responses effectively using simple regular expression matching. To address this issue, we design a post-processing program that leverages a dedicated language model to simplify and standardize outputs before scoring them with a regex-based approach.  

The post-processing program simplifies complex generated text, enabling it to conform to standardized scoring requirements. Some generated responses contain lengthy explanations, which complicates direct regex matching. The dedicated language model simplifies and reformats these outputs to enable subsequent scoring automation.

\textbf{Example}:
``Ground Truth Answer": ``A, B".
``LLM Response": ``Based on the text provided, the sentiment for 'Midwest' can be inferred as A. Negative, due to the implication of disruption caused by the travel advisory. For 'FAA', since it is an authoritative body issuing the advisory, the sentiment could be interpreted as B. Neutral, as it is not expressing a positive or negative opinion but rather providing factual information."

In this example, the ``LLM Response" output contains complex explanatory text, which makes it difficult to match directly with ``Ground Truth Answer" using regular expressions. With post-processing, however, the ``LLM Response" output is reformatted to \texttt{(A, B)}, allowing the regex-based scoring program to interpret and score the response accurately.

This post-processing strategy improves scoring accuracy and reduces the need for extensive formatting instructions in multimodal sentiment analysis, minimizing potential interference with model performance. We select the Qwen2.5-7B-Instruct model for this post-processing, as it demonstrates strong capabilities in instruction-following and structured output generation. Other language models with structured output capabilities are also suitable for this purpose.

\subsection{Metrics}

In the Benchmark evaluation study, a novel scoring method is introduced to enhance the assessment of MOSABench performance. Specifically, each question is designed as a multiple-choice task where each sample requires separate emotion assessments for object1 and object2. Following an examination-style grading approach, if both assessments are correct, the sample receives a score of 3; if only one assessment is correct, it receives a score of 1; and if both are incorrect, no points are awarded, with a maximum score of 3 for each sample.
To facilitate comparison with traditional benchmarks, each MLLM performance is also evaluated using standard metrics, including F1, Precision, and Recall. This combined assessment provides a comprehensive view of model effectiveness in multi-object sentiment analysis tasks.

\input{table/1_mllm_perf}

\input{table/2_score_perf}

\begin{figure*}[htbp]
    \centering
    \includegraphics[width=\linewidth]{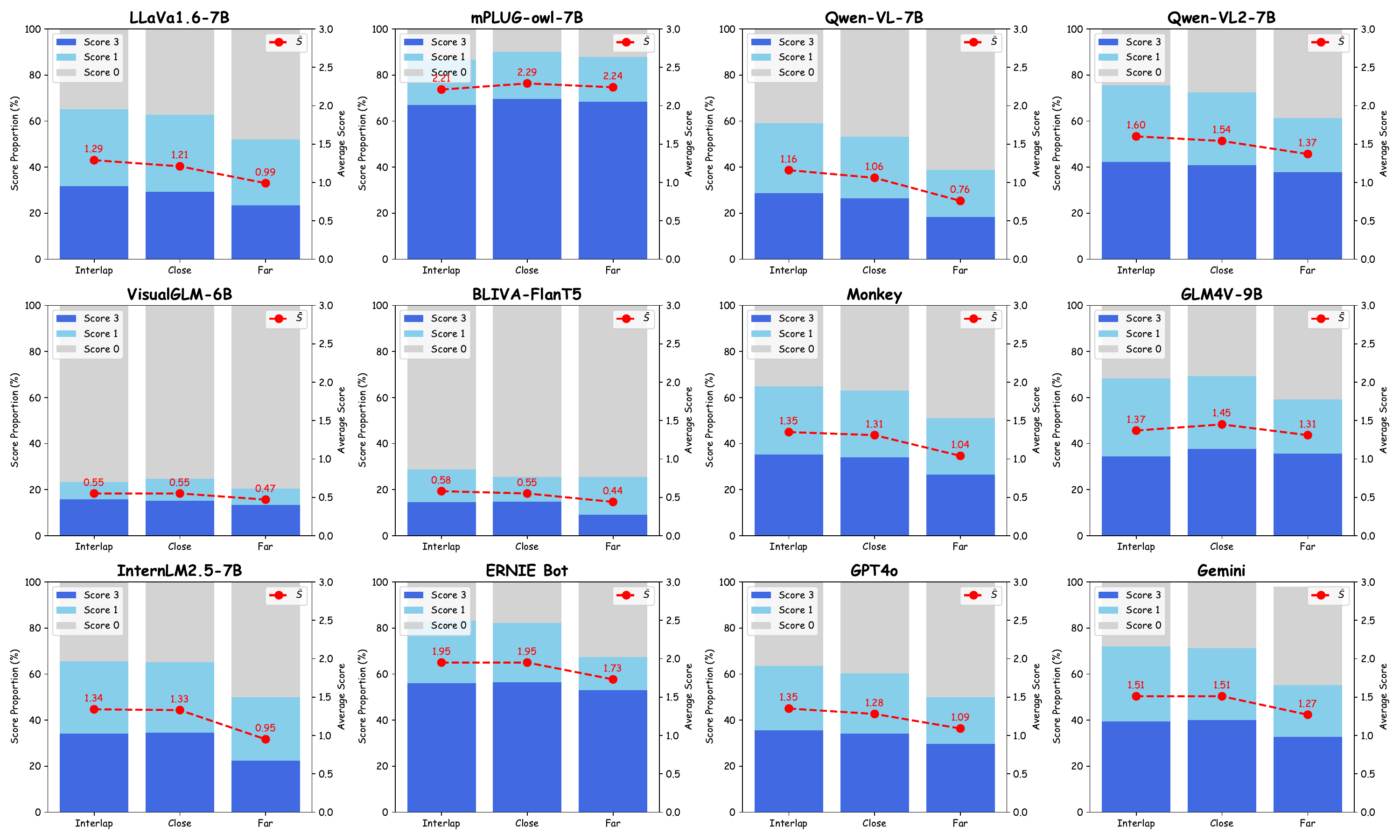}
    \caption{Analysis of MLLM Score Variations with Target Distance on MOSABench: average score $\bar{S}$ and distribution of $P_3$ (Correct), $P_1$ (Partially Correct), and $P_0$ (Incorrect) Scores.}
    \label{fig:S3_Proportion}
\end{figure*}

\begin{figure}[htbp]
    \centering
    \includegraphics[width=\linewidth]{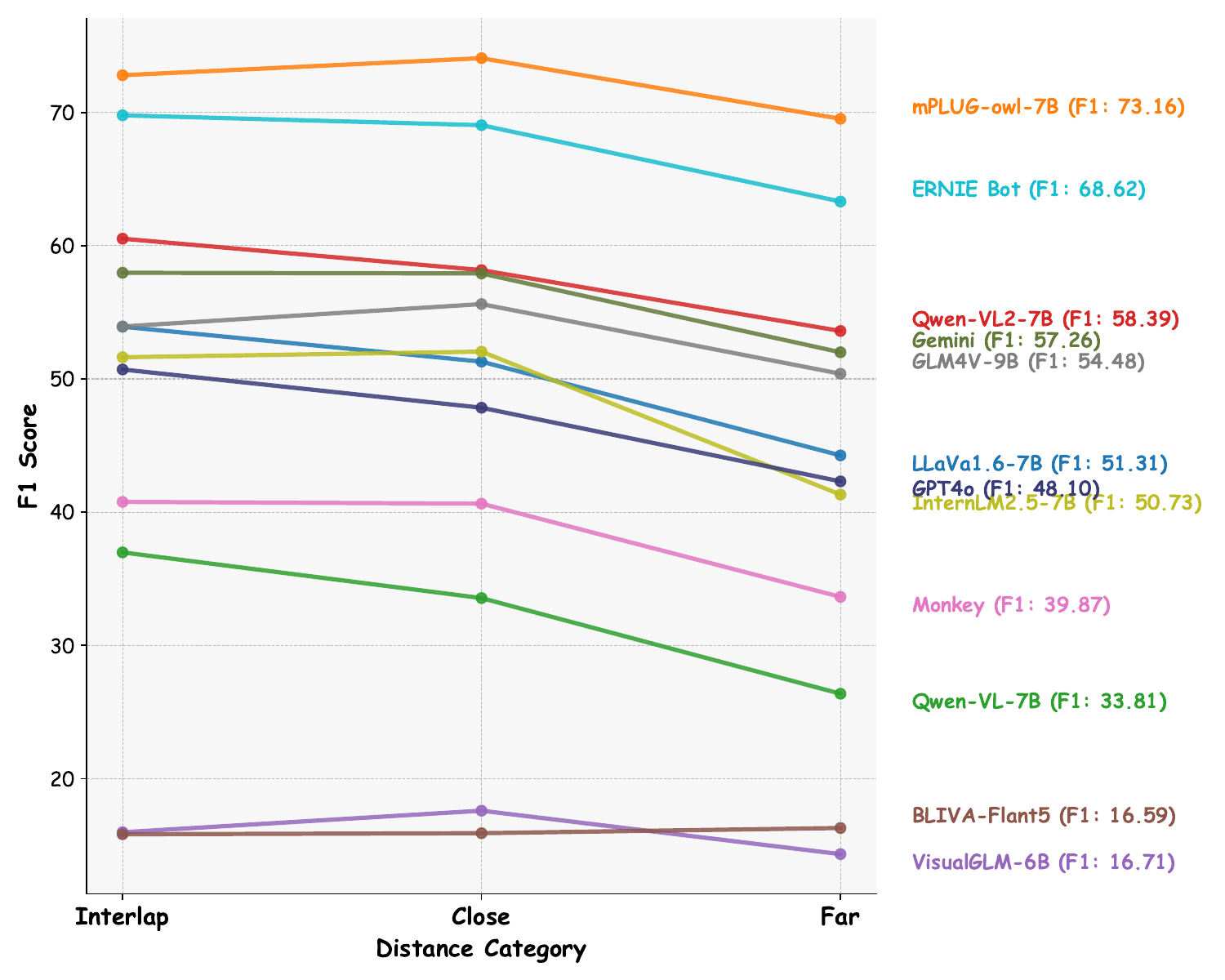}
    \caption{F1 across distance categories for different MLLMs}
    \label{fig:f1_line}
\end{figure}

\section{Experiments and Analysis}





\subsection{Main Result Analysis}
As shown in Table \ref{tab:mllm_performance}, we evaluate the performance of various MLLMs on MOSABench, focusing on multi-object sentiment analysis across different target distances. This table includes both open-sourced models (e.g., Qwen-VL, VisualGLM, Monkey) and close-sourced models (e.g., ERNIE Bot, GPT4o, Gemini), with performance F1, Precision, and Recall for three distance categories: Interlap, Close, and Far.
Open-sourced models exhibit a broad performance range. Among them, Qwen-VL2-7B achieves the highest overall F1 Score of 58.39, showing balanced performance across all categories. However, models like Qwen-VL-7B and VisualGLM-6B score significantly lower, especially with distant targets in the Far category. Close-sourced models generally outperform open-sourced ones, with ERNIE Bot achieving the highest overall F1 Score of 68.62, followed by Gemini at 57.26, indicating an advantage for close-sourced models in complex, multi-object sentiment analysis.

In the open-source category, Qwen-VL2-7B and GLM4V-9B demonstrate strong performance, particularly in the Close and Interlap categories, suggesting effective sentiment detection for nearby or overlapping targets. Monkey, with an overall F1 of 39.87, shows inconsistent performance, especially in the Far category, reflecting limitations with distant targets. Among closed-source models, ERNIE Bot consistently outperforms GPT4o and Gemini, maintaining high F1, Precision, and Recall scores, especially in Close and Interlap distances. GPT4o lags with an overall F1 score of 48.10, indicating challenges in multi-object sentiment analysis compared to ERNIE Bot and Gemini.

The results show that most models perform best in the Close category, followed by Interlap, with the lowest performance in the Far category. For example, ERNIE Bot scores an F1 of 69.79 in Interlap and 69.05 in Close, but drops to 63.32 in Far. This trend highlights the difficulty of interpreting emotional cues from distant objects, a challenge shared by both open-sourced and close-sourced models. The decline in metrics from Close to Far underscores the limitations of current MLLMs in identifying emotions across spatial resolutions.
The performance decline across distance categories and differences between models highlight the need for improvements in MLLM capabilities for multi-object sentiment analysis. While models like ERNIE Bot and mPLUG set high performance standards, especially for spatially proximate targets, low scores in the Far category reveal a gap in current architectures for handling distant emotional targets. Future research should focus on enhancing MLLM accuracy across diverse spatial contexts, particularly in challenging multi-object scenarios.

\begin{figure*}[htbp]
    \centering
    \begin{minipage}{0.15\linewidth}
        \centering
        \includegraphics[width=\linewidth]{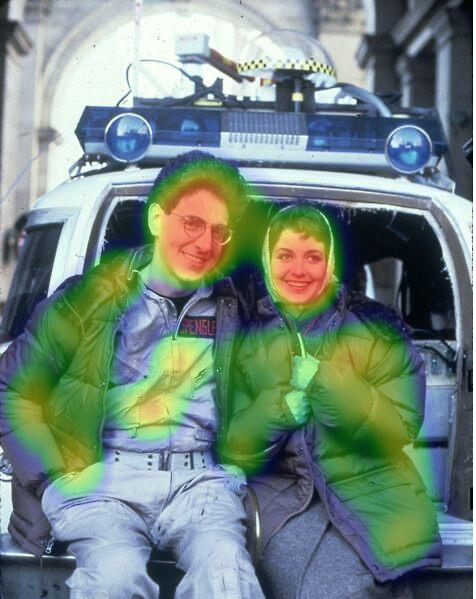}
        \caption*{mPLUG-Owl}  
        \label{fig:image1}
    \end{minipage}%
    \hspace{0.05\linewidth}
    \begin{minipage}{0.15\linewidth}
        \centering
        \includegraphics[width=\linewidth]{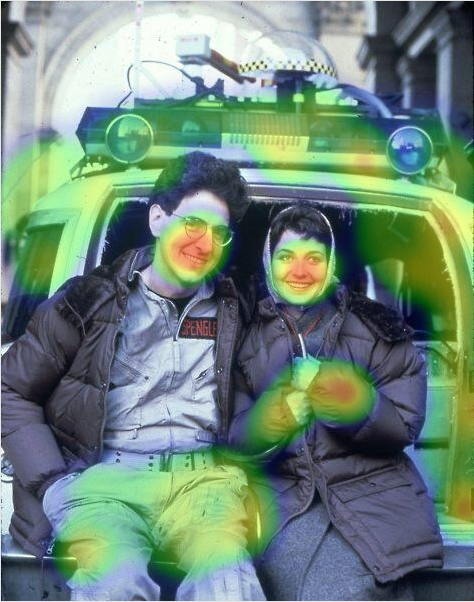}
        \caption*{Qwen-VL2}  
        \label{fig:image2}
    \end{minipage}%
    \hspace{0.05\linewidth}
    \begin{minipage}{0.15\linewidth}
        \centering
        \includegraphics[width=\linewidth]{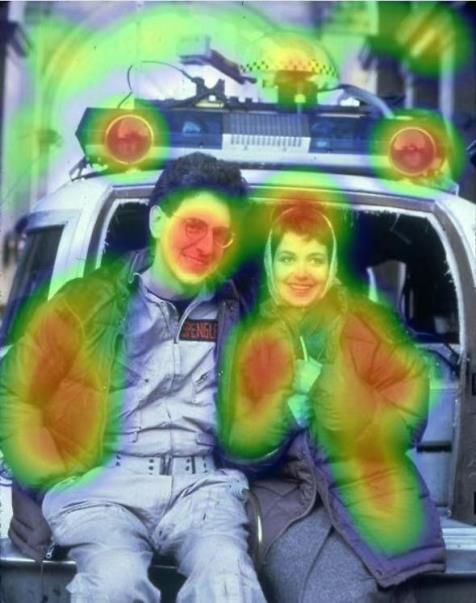}
        \caption*{LLaVA}  
        \label{fig:image3}
    \end{minipage}%
    \hspace{0.05\linewidth}
    \begin{minipage}{0.15\linewidth}
        \centering
        \includegraphics[width=\linewidth]{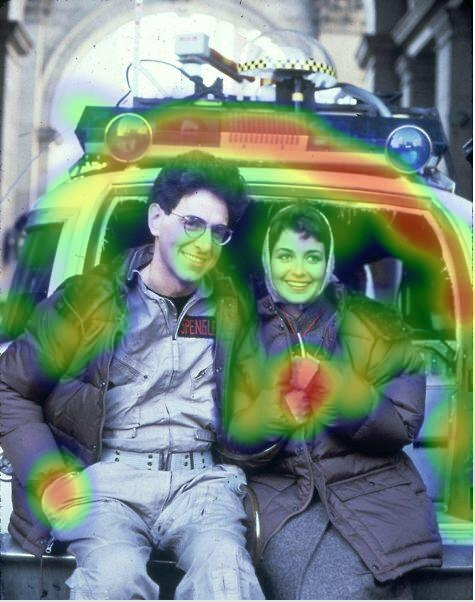}
        \caption*{VisualGLM}  
        \label{fig:image4}
    \end{minipage}
    \caption{Attention heatmaps of MLLMs. F1 scores: 73.16 (mPLUG-Owl), 58.39 (Qwen-VL2), 51.31 (LLaVA), and 16.71 (VisualGLM).}
    \label{fig:image_heatmap}
\end{figure*}

\begin{figure*}[htbp]
    \centering
    \begin{minipage}{0.23\linewidth}
        \centering
        \includegraphics[height=.8\linewidth]{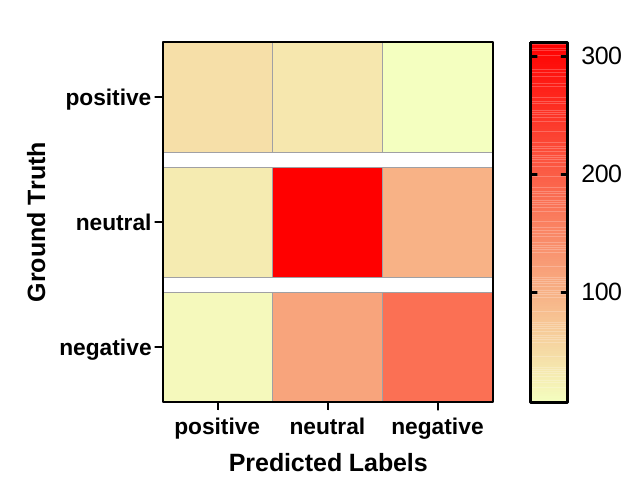}
        \vspace{0.1cm} 
        {\small \hspace{0.5cm}mPLUG-Owl}  
        \label{fig:image1}
    \end{minipage}%
    \hspace{0.02\linewidth}
    \begin{minipage}{0.23\linewidth}
        \centering
        \includegraphics[height=.8\linewidth]{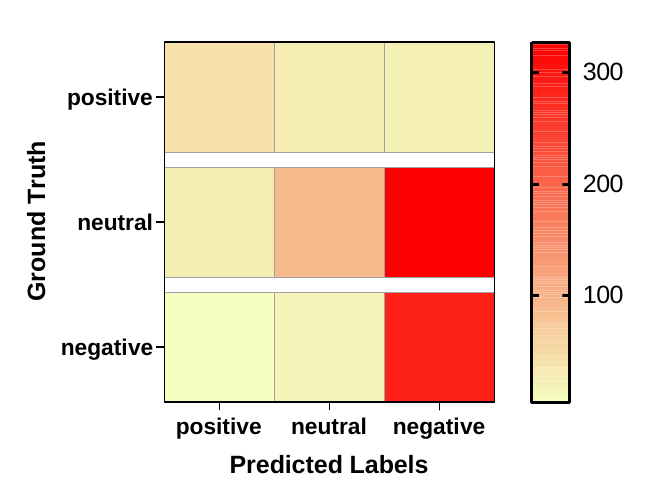}
        \vspace{0.1cm}
        {\small \hspace{0.5cm}Qwen-VL2}  
        \label{fig:image2}
    \end{minipage}%
    \hspace{0.02\linewidth}
    \begin{minipage}{0.23\linewidth}
        \centering
        \includegraphics[height=.8\linewidth]{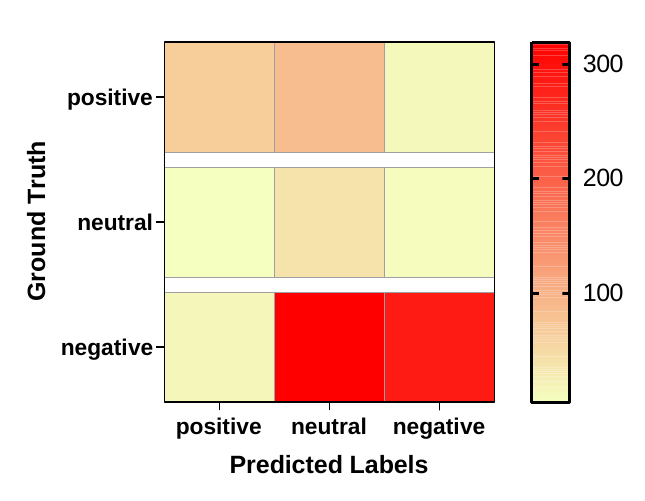}
        \vspace{0.1cm}
        {\small \hspace{0.5cm}LLaVA}  
        \label{fig:image3}
    \end{minipage}%
    \hspace{0.02\linewidth}
    \begin{minipage}{0.23\linewidth}
        \centering
        \includegraphics[height=.8\linewidth]{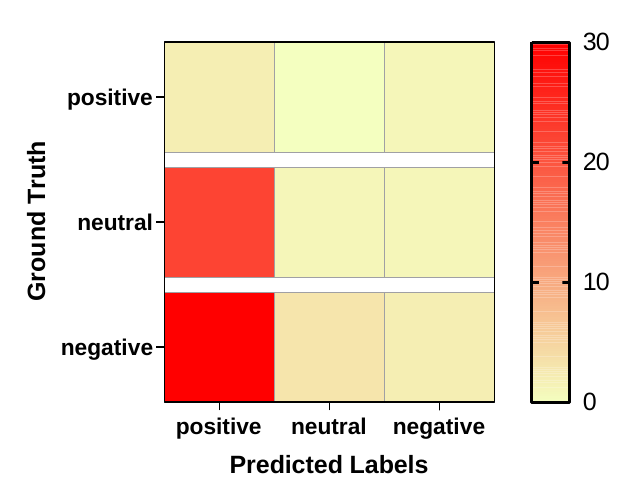}
        \vspace{0.1cm}
        {\small \hspace{0.5cm}VisualGLM}  
        \label{fig:image4}
    \end{minipage}
    \caption{The confusion matrix between actual predictions and ground truth of MLLMs}
    \label{fig:confusion_matrix}
\end{figure*}

\subsection{Score Distribution Across Distance Labels}
Table \ref{tab:score} shows model performance across Interlap, Close, and Far categories, highlighting a decline in correct predictions (S3) and average scores ($\bar{S}$) as object distance increases. This trend indicates that MLLMs struggle more with emotion recognition as targets become more distant, underscoring their limitations in complex, multi-object scenarios.
Among open-sourced models, Qwen-VL2-7B achieves the highest average score ($\bar{S}$) of 1.54, with strong S3 proportions across distances, showing effective emotion detection, especially for nearby objects. In contrast, Qwen-VL-7B performs the worst, with an average score of 0.31 and high S0 proportions, particularly in the Far category. ERNIE Bot leads among close-sourced models with a score of 1.93, maintaining robust performance across distances, while GPT4o, with a score of 1.28, struggles most in the Far category. This variability underscores the adaptability of models like ERNIE Bot and Qwen-VL2-7B, while others show limitations as distance increases.
Most models perform best in the Close category, followed by Interlap, and show their lowest performance in the Far category, highlighting shared challenges with distant targets. Notably, ERNIE Bot and Gemini maintain relatively high performance in the Far category, with $\bar{S}$ scores of 1.73 and 1.27, respectively, indicating greater robustness. Figure~\ref{fig:S3_Proportion} illustrates the decline in S3 scores and the increase in S0 scores as distance grows, particularly for models like Qwen-VL-7B and VisualGLM-6B.
The generally low scores in the Far category emphasize the need for improved MLLM architectures for accurate emotion detection in complex, multi-object scenarios. Future work should focus on enhancing MLLM capabilities to ensure consistent performance across varying spatial configurations.

\subsection{Attention Visualization Analysis}

Figure \ref{fig:image_heatmap} visualizes the attention regions of various MLLMs during multi-object sentiment analysis, highlighting how each model interprets visual information, particularly in recognizing facial expressions, which are crucial for accurate sentiment detection.
mPLUG-Owl, with the highest F1 (73.16), demonstrates the best focus, accurately targeting the facial expressions of both subjects. This focused attention suggests a strong understanding of task requirements, enabling it to avoid distractions and focus on sentiment-relevant features. Qwen-VL2, with an F1 of 58.39, also attends to the facial areas but with less intensity and consistency, indicating some missed or under-emphasized sentiment details, which may explain its lower accuracy compared to mPLUG-Owl.
LLaVA, with an F1 of 51.31, disperses its attention between faces and irrelevant areas, suggesting an inability to isolate sentiment-relevant regions effectively. This scattered focus likely introduces noise, reducing its accuracy. VisualGLM, with the lowest F1 (16.71), fails to target facial expressions and instead focuses on unrelated areas, significantly impairing its sentiment detection performance.
These visualizations underscore the importance of targeted attention in multi-object sentiment analysis: models that concentrate on sentiment-relevant areas, like facial expressions, achieve higher accuracy, while those with dispersed or misaligned focus, such as VisualGLM, perform poorly.

\subsection{Confusion Matrix Analysis}

Figure \ref{fig:confusion_matrix} presents the confusion matrices for four MLLMs, illustrating their performance in multi-object sentiment analysis, with redder cells indicating higher counts for each sentiment classification.
mPLUG-Owl, the model with the highest performance, shows strong results along the diagonal, with the reddest cells indicating it accurately classifies sentiments across categories, including neutral (neu) and negative (neg) sentiments. Qwen-VL2 also performs well but has a tendency to misclassify neutral sentiments as negative, as indicated by the red cells off the diagonal in the neutral-negative area. LLaVA, in contrast, often misclassifies negative sentiments as neutral, as shown by the reddish cells in the negative-neutral category. VisualGLM, the model with the weakest performance, struggles significantly, frequently misclassifying negative sentiments as positive, with noticeable red shading in the off-diagonal cells of the negative-positive category.
This analysis highlights the distinct ways each model handles sentiment classification errors, with mPLUG-Owl achieving the most accurate classifications, while VisualGLM exhibits considerable challenges in sentiment analysis.

\section{Conclusion}

We systematically investigate the limitations of MLLMs in multi-object sentiment analysis, particularly in scenarios with visually distinct and spatially varied targets. We introduce MOSABench, a benchmark dataset specifically designed to evaluate MLLM capabilities in independently and accurately assessing sentiments across multiple objects within a single image. MOSABench includes a wide range of spatial relationships, enabling a detailed analysis of how target proximity and feature diversity impact model performance.
Our findings reveal the significant challenges MLLMs face in complex multi-object environments, highlighting the need for architectural enhancements to improve their adaptability to these tasks. By providing a dedicated dataset and comprehensive evaluation framework, this work lays a foundation for future research aimed at advancing MLLM performance in nuanced, multi-target sentiment analysis.
MOSABench serves as an initial exploration for future MLLM research, offering directions and insights for evaluating multi-object sentiment analysis. This benchmark aims to contribute to the development and assessment of models better suited for complex multimodal tasks, supporting progress in multi-object sentiment understanding.

\bibliography{main}
\bibliographystyle{IEEEtranN}

\end{document}

%% file: table/3_alg.tex
\begin{algorithm}[h]
    \caption{Calculate Distance Type Between Two Bounding Boxes}
    \label{alg:distance_type}
    \begin{algorithmic}[1]
    \Require Image length $L$, Bounding boxes $\mathbf{B_1} = (x_1, y_1, x_1', y_1')$ and $\mathbf{B_2} = (x_2, y_2, x_2', y_2')$, threshold parameter $k$
    \Ensure Distance label: \textit{Interlap}, \textit{Close}, or \textit{Far}
    
    \Statex \textbf{Step 1: Compute center points} $\mathbf{C_1}$ and $\mathbf{C_2}$:
    \Statex \hspace{1em} $C_1 = \left( \frac{x_1 + x_1'}{2}, \frac{y_1 + y_1'}{2} \right)$
    \Statex \hspace{1em} $C_2 = \left( \frac{x_2 + x_2'}{2}, \frac{y_2 + y_2'}{2} \right)$
    \Comment{Calculate the midpoint of each bounding box}
    
    \Statex \textbf{Step 2: Check for overlap} (\textit{Interlap})
    \If{$x_1 < x_2'$ \textbf{and} $x_1' > x_2$ \textbf{and} $y_1 < y_2'$ \textbf{and} $y_1' > y_2$}
        \State \Return \textit{Interlap} \Comment{Bounding boxes overlap}
    \EndIf
    
    \vspace{0.5em}
    \Statex \textbf{Step 3: Calculate the Euclidean distance} $d$ \textbf{between} $C_1$ \textbf{and} $C_2$:
    \Statex \hspace{1em} $d = \sqrt{(C_{1x} - C_{2x})^2 + (C_{1y} - C_{2y})^2}$
    \Comment{Distance between centers of $B_1$ and $B_2$}
    
    \vspace{0.5em}
    \Statex \textbf{Step 4: Determine distance type} (\textit{Close} or \textit{Far}):
    \If{$d < L/k$}
        \State \Return \textit{Close} \Comment{Boxes are close to each other}
    \Else
        \State \Return \textit{Far} \Comment{Boxes are far from each other}
    \EndIf
    
    \end{algorithmic}
\end{algorithm}

%% file: table/data_statistics.tex
\begin{table}[htbp]
\centering
\small 
\resizebox{\linewidth}{!}{
    \begin{tabular}{|l|l|}
    \hline
    \textbf{Total Samples} & 1047 \\ \hline
    \textbf{Shortest Text Length} & 20 \\ \hline
    \textbf{Longest Text Length} & 146 \\ \hline
    \textbf{Average Text Length} & 86.27 \\ \hline
    \textbf{Answer Proportions (Neg, Neu, Pos)} & 15.4\%, 49.4\%, 35.1\% \\ \hline
    \textbf{Distance Proportions (I, C, F)} & 31.2\%, 57.1\%, 11.7\% \\ \hline
    \end{tabular}
    }
\caption{Statistical overview of MOABench. I, C, and F represent the distance categories: Interlap, Close, and Far, respectively.}
\label{tab:dataset_statistics}
\end{table}

%% file: table/1_mllm_perf.tex
\begin{table*}[htbp]
\centering
\caption{F1, precision and recall comparison of MLLMs on MOSABench across various objects distances (Interlap, Close, Far).}
\resizebox{\textwidth}{!}{ 
    \begin{tabular}{l|ccc|ccc|ccc|ccc}
    \toprule
    \multicolumn{13}{c}{\textbf{Open-sourced Models}} \\
    \midrule
    \multicolumn{1}{c|}{\multirow{2}[2]{*}{\textbf{MLLM}}} & \multicolumn{3}{c|}{\textbf{Overall}} & \multicolumn{3}{c|}{\textbf{Interlap}} & \multicolumn{3}{c|}{\textbf{Close}} & \multicolumn{3}{c}{\textbf{Far}} \\
          & \textbf{F1} & \textbf{Precision} & \textbf{Recall} & \textbf{F1} & \textbf{Precision} & \textbf{Recall} & \textbf{F1} & \textbf{Precision} & \textbf{Recall} & \textbf{F1} & \textbf{Precision} & \textbf{Recall} \\
    \midrule
    \textbf{LLaVA1.6-7B\cite{llava}} & 51.31  & 51.31  & 46.87  & 53.92  & 59.08  & 49.58  & 51.30  & 57.30  & 46.43  & 44.26  & 47.37  & 41.54  \\
    \textbf{mPLUG-owl-7B\cite{mPLUG-Owl}} & \textbf{73.16} & \textbf{69.86} & \textbf{76.78} & \textbf{72.80} & \textbf{70.18} & \textbf{75.62} & \textbf{74.08} & \textbf{70.66} & \textbf{77.85} & \textbf{69.53} & \textbf{65.10} & \textbf{74.62} \\
    \textbf{Qwen-VL-7B\cite{qwen_vl}} & 33.81  & 29.14  & 40.26  & 36.98  & 31.86  & 44.04  & 33.55  & 28.93  & 39.91  & 26.37  & 22.65  & 31.54  \\
    \textbf{Qwen-VL2-7B\cite{Qwen2VL}} & 58.39  & 61.72  & 55.39  & 60.53  & 64.09  & 57.34  & 58.17  & 61.63  & 55.08  & 53.60  & 55.83  & 51.54  \\
    \textbf{VisualGLM-6B\cite{du2022glm}} & 16.71  & 14.22  & 20.26  & 15.97  & 13.72  & 19.11  & 17.59  & 14.94  & 21.40  & 14.33  & 12.04  & 17.69  \\
    \textbf{BLIVA-Flant5\cite{BLIVA}} & 16.59  & 14.54  & 19.30  & 15.82  & 20.78  & 17.96  & 15.90  & 13.93  & 18.51  & 16.29  & 14.12  & 19.23  \\
    \textbf{Monkey\cite{li2024monkey}} & 39.87  & 33.63  & 48.96  & 40.77  & 34.48  & 49.86  & 40.64  & 34.35  & 49.77  & 33.64  & 27.92  & 42.31  \\
    \textbf{GLM4V-9B\cite{CogVLM}} & 54.48  & 57.10  & 52.09  & 53.94  & 56.92  & 51.25  & 55.62  & 58.56  & 52.96  & 50.39  & 50.78  & 50.00  \\
    \textbf{InternLM2.5-7B\cite{InternLM2TechiReport}} & 50.73  & 53.36  & 48.35  & 51.63  & 52.91  & 50.42  & 52.05  & 55.29  & 49.17  & 41.32  & 44.64  & 38.46  \\
    \midrule
    \multicolumn{13}{c}{\textbf{Close-sourced Models}} \\
    \midrule
    \multicolumn{1}{c|}{\multirow{2}[2]{*}{\textbf{MLLM}}} & \multicolumn{3}{c|}{\textbf{Overall}} & \multicolumn{3}{c|}{\textbf{Interlap}} & \multicolumn{3}{c|}{\textbf{Close}} & \multicolumn{3}{c}{\textbf{Far}} \\
          & \textbf{F1} & \textbf{Precision} & \textbf{Recall} & \textbf{F1} & \textbf{Precision} & \textbf{Recall} & \textbf{F1} & \textbf{Precision} & \textbf{Recall} & \textbf{F1} & \textbf{Precision} & \textbf{Recall} \\
    \midrule
    \textbf{ERNIE Bot\cite{sun2020ernie}} & \textbf{68.62} & \textbf{70.18} & \textbf{67.13} & \textbf{69.79} & \textbf{71.51} & \textbf{68.14} & \textbf{69.05} & \textbf{70.81} & \textbf{67.37} & \textbf{63.32} & \textbf{63.57} & \textbf{63.08} \\
    \textbf{GPT4o\cite{GPT4}} & 48.10  & 48.92  & 47.30  & 50.71  & 51.88  & 49.58  & 47.84  & 48.67  & 47.04  & 42.31  & 42.31  & 42.31  \\
    \textbf{Gemini\cite{geminiteam2024geminifamilyhighlycapable}} & 57.26  & 60.29  & 54.52  & 57.97  & 60.79  & 55.40  & 57.92  & 61.25  & 54.93  & 52.00  & 54.17  & 50.00  \\
    \bottomrule
    \end{tabular}%
}
\label{tab:mllm_performance}
\end{table*}

%% file: table/2_score_perf.tex
\begin{table*}[htbp]
\centering
\caption{Score comparison of MLLMs on MOSABench across various objects distances (Interlap, Close, Far). ($\bar{S}$) is the average score. $P_0$, $P_1 $, and $P_3$ denote the proportions of Incorrect(0), Partial Correct(1), and Full Correct(3) samples, respectively.}
\resizebox{\textwidth}{!}{ 
    \begin{tabular}{l|cccc|cccc|cccc|cccc}
    \toprule
    \multicolumn{17}{c}{\textbf{Open-sourced Models}} \\
    \midrule
    \multicolumn{1}{c|}{\multirow{2}[2]{*}{\textbf{MLLM}}} & \multicolumn{4}{c|}{\textbf{Overall}} & \multicolumn{4}{c|}{\textbf{Interlap}} & \multicolumn{4}{c|}{\textbf{Close}} & \multicolumn{4}{c}{\textbf{Far}} \\
          & $P_3$(\%) & $P_1$(\%) & $P_0$(\%) &$\bar{S}$ & $P_3$(\%) & $P_1$(\%) & $P_0$(\%) &$\bar{S}$ & $P_3$(\%) & $P_1$(\%) & $P_0$(\%) &$\bar{S}$ & $P_3$(\%) & $P_1$(\%) & $P_0$(\%) &$\bar{S}$ \\
    \midrule
    \textbf{LLaVA1.6-7B\cite{llava}} & 29.39  & 32.86  & 37.75  & 1.21  & 31.80  & 33.33  & 34.87  & 1.29  & 29.29  & 33.47  & 37.24  & 1.21  & 23.47  & 28.57  & 47.96  & 0.99  \\
    \textbf{mPLUG-owl-7B\cite{mPLUG-Owl}} & 68.70  & 20.07  & 11.23  & \textbf{2.26} & 67.05  & 19.54  & 13.41  & \textbf{2.21} & 69.67  & 20.50  & 9.83  & \textbf{2.29} & 68.37  & 19.39  & 12.24  & \textbf{2.24} \\
    \textbf{Qwen-VL-7B\cite{qwen_vl}} & 26.16  & 27.12  & 46.71  & 1.06  & 28.74  & 30.27  & 41.00  & 1.16  & 26.36  & 26.78  & 46.86  & 1.06  & 18.37  & 20.41  & 61.22  & 0.76  \\
    \textbf{Qwen-VL2-7B\cite{Qwen2VL}} & 40.86  & 31.18  & 27.96  & 1.54  & 42.15  & 33.33  & 24.52  & 1.60  & 40.79  & 31.59  & 27.62  & 1.54  & 37.76  & 23.47  & 38.78  & 1.37  \\
    \textbf{VisualGLM-6B\cite{du2022glm}} & 15.17  & 8.60  & 76.22  & 0.54  & 15.71  & 7.66  & 76.63  & 0.55  & 15.27  & 9.41  & 75.31  & 0.55  & 13.27  & 7.14  & 79.59  & 0.47  \\
    \textbf{BLIVA-Flant5\cite{BLIVA}} & 14.10  & 12.43  & 73.48  & 0.55  & 14.56  & 14.18  & 71.26  & 0.58  & 14.85  & 10.67  & 74.48  & 0.55  & 9.18  & 16.33  & 74.49  & 0.44  \\
    \textbf{Monkey\cite{li2024monkey}} & 33.57  & 28.55  & 37.87  & 1.29  & 35.25  & 29.50  & 35.25  & 1.35  & 34.10  & 28.87  & 37.03  & 1.31  & 26.53  & 24.49  & 48.98  & 1.04  \\
    \textbf{GLM4V-9B\cite{CogVLM}} & 36.44  & 31.30  & 32.26  & 1.41  & 34.48  & 33.72  & 31.80  & 1.37  & 37.66  & 31.59  & 30.75  & 1.45  & 35.71  & 23.47  & 40.82  & 1.31  \\
    \textbf{InternLM2.5-7B\cite{InternLM2TechiReport}} & 32.97  & 30.47  & 36.56  & 1.29  & 34.10  & 31.42  & 34.48  & 1.34  & 34.52  & 30.54  & 34.94  & 1.33  & 22.45  & 27.55  & 50.00  & 0.95  \\
    \midrule
    \multicolumn{17}{c}{\textbf{Close-sourced Models}} \\
    \midrule
    \multicolumn{1}{c|}{\multirow{2}[2]{*}{\textbf{MLLM}}} & \multicolumn{4}{c|}{\textbf{Overall}} & \multicolumn{4}{c|}{\textbf{Interlap}} & \multicolumn{4}{c|}{\textbf{Close}} & \multicolumn{4}{c}{\textbf{Far}} \\
          & $P_3$(\%) & $P_1$(\%) & $P_0$(\%) &$\bar{S}$ & $P_3$(\%) & $P_1$(\%) & $P_0$(\%) &$\bar{S}$ & $P_3$(\%) & $P_1$(\%) & $P_0$(\%) &$\bar{S}$ & $P_3$(\%) & $P_1$(\%) & $P_0$(\%) &$\bar{S}$ \\
    \midrule
    \textbf{ERNIE Bot\cite{sun2020ernie}} & 55.91  & 24.85  & 19.24  & \textbf{1.93} & 55.94  & 27.20  & 16.86  & \textbf{1.95} & 56.49  & 25.73  & 17.78  & \textbf{1.95} & 53.06  & 14.29  & 32.65  & \textbf{1.73} \\
    \textbf{GPT4o\cite{GPT4}} & 34.05  & 26.05  & 39.90  & 1.28  & 35.63  & 27.97  & 36.40  & 1.35  & 34.10  & 26.15  & 39.75  & 1.28  & 29.59  & 20.41  & 50.00  & 1.09  \\
    \textbf{Gemini\cite{geminiteam2024geminifamilyhighlycapable}} & 39.19  & 30.59  & 30.23  & 1.48  & 39.46  & 32.57  & 27.97  & 1.51  & 39.96  & 31.17  & 28.87  & 1.51  & 32.69  & 22.45  & 42.86  & 1.27  \\
    \bottomrule
    \end{tabular}%
}
\label{tab:score}%
\end{table*}%